# Rule based Part of speech Tagger for Homoeopathy Clinical realm

Sanjay K. Dwivedi[1] and Pramod P. Sukhadeve[3]

[1] Department of Computer Science, Babasaheb Bhimrao Ambedkar University
Lucknow, Uttar Pradesh 226025, India

[2] Department of Computer Science, Babasaheb Bhimrao Ambedkar University
Lucknow, Uttar Pradesh 226025, India

**Abstract**
A tagger is a mandatory segment of most text scrutiny systems, as it consigned a s yntax class (e.g., noun, verb, adjective, and adverb) to every word in a sentence. In this paper, we present a simple part of speech tagger for homoeopathy clinical language. This paper reports about the anticipated part of speech tagger for homoeopathy clinical language. It exploit standard pattern for evaluating sentences, untagged clinical corpus of 20085 words is used, from which we had selected 125 sentences (2322 tokens). The problem of tagging in natural language processing is to find a way to tag every word in a text as a meticulous part of speech. The basic idea is to apply a set of rules on clinical sentences and on each word, Accuracy is the leading factor in evaluating any POS tagger so the accuracy of proposed tagger is also conversed.
Keywords: **POS tagging, Natural language processing, Grammar rules, Homoeopathic Corpus..**

## 1. Introduction

Part of speech tagging is a process of assigning accurate syntactic categories to every word in the text as corresponding to a particular part of speech, based on its definition, as well as its context [1]. POS tagging is a very important pre-processing task for language processing performance. This facilitates in doing profound parsing of text and in developing information extraction systems, semantic processing etc. POS tagging for natural language texts have been developed via linguistic rules. It plays elementary role in various Natural Language Processing applications such as speech recognition extraction; machine translation and word sense disambiguation etc. although POS tagging for clinical language has gained an increased interest over the past few years, yet the lack of availability of annotated corpora resources obstruct the research and investigations, standardization is another problem because so far no standard tag sets are available for such languages. While so far this is the situation for homoeopathy clinical languages. Much medical information subsists as free-from text, from patient histories, through discharge summaries, to journal articles detailing new discoveries and information about participation in clinical trials.

## 2. Overview of POS Tagging

Every language has its parts of speech for instance verb, noun, adjective…etc. POS tagging is a p rocedure of spontaneous allocate the POS for the word. It is raised area on its definition, in addition to its context. It can be exploit for text parsing, information extraction, text review and machine translation. There are convinced approaches like stochastic approach [2, 3] uses a t raining corpus to acknowledge the most credible tag for a word. Part-of-speech (POS) tagging is universally known as the assignment of categorizing a word in a s pecified input sentence by allocating it a t ag from a p redefined set of module that symbolize syntactic behaviour. For languages such as English, word-level POS tagging appears sufficient because words typically correspond to the syntactically pertinent POS tag classes. The first step in building a part of speech tagger is to assemble a lexicon, where the part of speech of a w ord can be initiated. Unfortunately many words are ambiguous and each word can consequently have several classifications. As an example "patient" can be either an adjective or a noun. It is the objective of the part of speech tagger to determine these ambiguities, using the scaffold of the words. Another example that taggers face ambiguity, even a w ord occurs in a l exicon, it may have many senses or meanings. A common example from the medical domain is "dose." "Dose' can be a noun, meaning the amount of medicine the patient should take, or it can be a verb, meaning the activity of giving medication to a patient.

There are principally two approaches to part-of-speech tagging: rule based tagging and stochastic tagging. This paper describes a rule based approach. Some of the tag sets use for Rule based clinical POS tagger is as follows,

Table 1: Part of Speech Tag sets

| Tag | Description | Tag | Description |
|-----|-------------|-----|-------------|
| ADV | Adverb | PO | Ordinal pronoun |
| AVB | Adverbial particle | PP | Personal pronoun |
| CND | Conditional | PPI | Inflectional post position |
| CNJ | Conjunction | PPP | Possessive post position |
| ADJ | Adjective | PQ | Question marker |
| DTR | Relative Determiner | PPH | Preposition |
| ETC | Conuation Marke | PT | Temporal pronoun |
| FW | Foreign Word | QUA | Qualifier |
| INT | Interjection | RPP | Personal relative pronoun |
| JF | Following Adjectives | RPS | Spatial relative pronoun |
| JJ | Noun Qual. Adjectives | RPT | Temporal relative pronoun |
| JQC | Cardinal Qual. adjectives | SEN | Sentinel |
| INPR | Interogative Pronoun | SHD | Semantic Shades incurring particle |
| JQQ | Quantifier | SYM | Symbol |
| NEG | Negative | LVB | Linking Verb |
| NN | Common Noun | VF | Finite Verb |
| NP | Proper Noun | VIS | Imperative Verbs |
| NUM | Number | VM | Modal Verb |
| NV | Verbal noun | VN | Non-Finite Verb |
| PC | Cardinal pronoun | VNG | Verb Negative |

Above table shows the complete tag sets used in Homoeopathy tagger. Although we delineate 40 tags for proposed Homoeopathy POS tagger, several of them are still misplaced which may necessitate further research and development.

## 3. Related Work

Special approaches have been used for Part-of-Speech (POS) tagging, where the prominent solitaries are rule-based, stochastic, or transformation-based learning approaches. The stochastic (probabilistic) approach [4, 5] uses a training corpus to accepted nearly all credible tag for a word. All probabilistic methods cited above are based on first order or second order Markov models. There are a few other methods which use probabilistic approach for POS tagging, such as the Tree Tagger [6]. Lastly, the transformation-based loom combines the rule-based approach and statistical approach. It selects the most likely tag based on a training corpus and then pertain a persuaded set of rules to see whether the tag should be changed to anything else. It saves any new rules that it has learnt in the development, for future use. One of the effective tagger is the Brill tagger [7, 8]. Rule-based taggers [9] try to allocate a tag to each word using a set of hand-written rules. These rules could stipulated, for instance, that a verb follows 'to' it is an infinitive phrasing not the main verb. You will find the main verb either before or after the infinitive phrase. Of course, this means that the set of rules must be appropriately written and inveterate by human experts. In existence there are three taggers, dTaggers [10], MaxEnt[11] and Curran & Clark[12]. Out of which dTagger is used for clinical texts and other two for news and articles. These taggers have relatively analogous rates of accuracy: dTagger, MaxEnt, and Curran & Clark had 87%, 89%, and 90 % respectively.

## 4. Proposed Technique and their rules

Various techniques have been explored for Part-of-Speech tagging [13]. Some of these are entirely automated while others necessitate a lot of human input. The primary step towards development of a Rule Based Part-of-Speech tagger for any language demands an in-depth understanding and analysis of that language [14]. primarily, we perceive certain rules for sentence analysis and tagging each word duly. Following figure shows the steps for analyzing. No matter how long a sentence is or how difficult a sentence appears to be, analyzing the sentence is easy now by using the simple steps. It will be easier to analyze more complicated sentences.

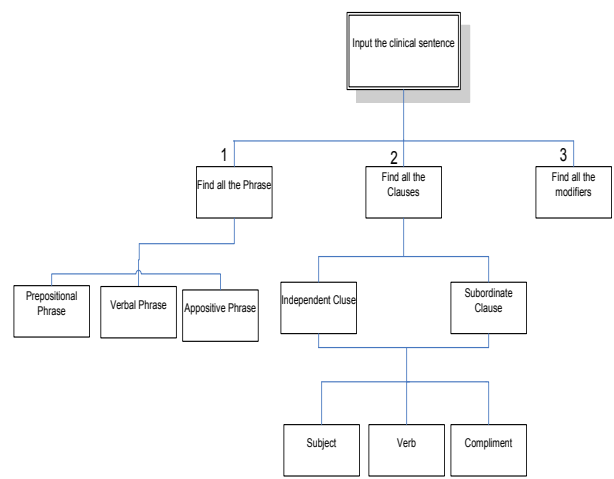

Fig. 1 Detail steps of sentence analysis.

Step 1- Scan the entire sentence and looking for all the phrases (Prepositional Phrase, Verbal Phrase and appositive Phrase) and tag all the modifiers suitably by applying the pertinent rules.

Step 2- Scan the remaining sentence to identify the core of all clauses. The easiest way to do this is to find all the verbs first and then identify the subjects, verbs and compliments, tag all the words by applying the pertinent rules.

Step 3- Find the remaining modifiers by using relevant rules and tag them pertinently.

4.1 Depiction of each steps

Initially we contribute sentence of any type then firstly we analyze the sentence for concerning the grammatical rules to each words. There are total 485 grammar rules in all. Category wise these rules are as follows, noun and noun clause has 90 rules, verb and verbal phrase has 102 rules, adjective and adjective clause has 77 rules preposition has 38 rules. Punctuation has 98 rules. Adverb, adverb clause and modifiers have 21, 13 and 46 rules respectively. Some of the rules are enlightened in each step such as,

Step1. Find all the Phrases

Prepositional Phrase - Preposition Phrase commences with a preposition and ends with a noun or pronoun. Whereas noun or pronoun is called the object of preposition, Subject and Verbs are never found within the prepositional phrase. Only Adjectives and Adverbs are present in prepositional phrase. After identifying the prepositional phrase and modifiers in it tag them relevantly.

Verbal Phrase - A verbal phrase consists of a verbal and all of its modifiers and objects. Since verbal come from verbs, a verbal phrase consists of a verbal and all of its modifiers and objects. Since verbal come from verbs.

Appositive Phrase - An appositive is a noun or noun phrase that renames another noun right adjacent to it. The appositive can be a short or long combination of words. The imperative point to remember is that a nonessential appositive is always separated from the respite of the sentence with comma(s).

Step 2. Find all the Clauses

Independent Clause - Independent clause in one that composes sense standing alone (a simple sentence).

Subordinate Clause - A subordinate clause does not make a complete sense, it may be used in three ways in a sentence: as an adjective, as an adverb, or as a noun.

The Adjective clause:- used as an adjective to modify a noun or pronoun. It begins with a relative pronoun or with the adverbs, "when" and "where." An adjective clause usually modifies the noun or pronoun that immediately precedes it. Therefore, an adjective clause will never be found at the beginning of the sentence. An adjective clause will contain a subject and verb and any other element that can be found in a sentence. Often, the relative pronoun is one of the important elements in the clause.

The Adverb clause:- used as an adverb and usually modifies the verb in the independent clause. It begins with a subordinate conjunction or with an adverb. When it appears at the beginning of the sentence should be followed by a comma. An adverb clause will contain a subject and verb and any other element that can be found in a sentence.

The Noun clause:- used as an noun and can be the subject, direct object, predicate nominative, or object of the preposition in the sentence. It usually begins with a relative pronoun. It may contain a subject and verb and any other ingredient that can be found in a sentence. To determine the function of a noun clause, first see if there is a preposition in front of it. If there isn't, look at the position of the clause in the sentence. If the clause is at the beginning of the sentence, it is the subject. If it is at the end of the sentence, it will be a direct object or a predicate nominative.

Step 3.Find all the modifiers by applying relevant rules

In the last step we will find all the modifiers by applying relevant rules, some of the rules of modifiers are mention below,

a) Modifier - Describes/indentifies someone or something else in the sentence
b) Be on the lookout of the opening modifiers, which appear at the beginning of the sentence
c) The opening modifiers modify the nouns that follow them.
d) The opening modifiers are separated from the rest of the sentence by comma. The sentence contains the noun being modified.
e) Adjectives modifies noun or pronoun
f) Adverb modifies verb, adjective, another adverb, a propositional phrase, or even a whole clause.
g) Adverb cannot modify noun or pronoun
h) Adverbs are formed by adding "ly" to the adjective.
i) Adjectives, not adverbs, follow the linking verbs such as feel, seem. These adjectives do not modify the verb but indentify a quality with a noun subject. Linking verbs illustrate what the subject is or what condition the subject is in and not what action the subject is doing.

## 5. Proposed architecture for POS Tagging

The connotation of part of speech for language processing is the large amount of information they give about a word and its neighbors. The proposed tag set for clinical English Language has 40 tags. The proposed architecture for POS tagging is shown below:

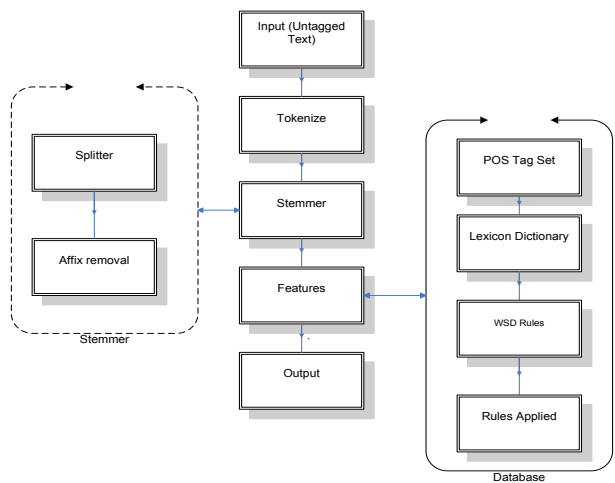

Fig. 2 Architechture of POS Tgger.

The POS tagging architecture consists of different modules which achieve different functionalities to accomplish better accuracy of POS tagger. Firstly we input the untagged text then tokenize it, after tokenizing it selects the single words for splitting and remove its affix by using the stemmer, and then it moves to the features in which the particular words are grammatically categories by using lexicon Dictionary, POS Tag set and grammar rules. The proposed clinical tagger is totally domain specific; in this technique we have not enlightened the word sense disambiguation part which may require further research and development. One of the WSD problems that taggers face is ambiguity. Even if a word occurs in a lexicon, it may have many senses or meanings. A common example from the medical domain is "dose." "Dose' can be a noun, meaning the amount of medicine the patient should take, or it can be a verb, meaning the activity of giving medication to a patient.

## 6. Result Anakysis

The precision of any part of speech tagger is measured in terms of percentage i.e. the percentage of words, which are accurately tagged by the tagger. This is defined as below

$$Accuracy = \frac{CorrectlyTaggedWords}{TotalNo.ofNumberTagged}$$

For evaluating proposed tagger, a corpus having text from special homoeopathy books [15], medical reports, symptoms [16] and prescriptions [17]. The outcome was manually appraised to mark the correct and incorrect tag assignments. 125 sentences (2322 words) collected randomly from 20085 words corpus of homoeopathy were manually appraised and are grouped into four different diseases. Only four diseases are to be taken from the complete corpus for tagging.

Table 2: Performance of Part of Speech Tagger

| *Corpus Diseases* | *Tagged Words* | | *Total Words* |
|---|---|---|---|
| | *Incorrect Tag* | *Correct Tag* | |
| Rheumatism | 28 | 421 | 449 |
| Anaemia | 58 | 735 | 793 |
| Migraine | 30 | 130 | 160 |
| Keloids | 110 | 810 | 920 |
| Total | 226 | 2096 | 2322 |

Table 2 shows the performance of part of speech tagger, sentences are collected from the manually built clinical (homoeopathy) corpus. We acquired sentences from some of the diseases like Rheumatism, Anaemia, Migraine, Keloids. Correctly tagged words from Rheumatism are 421 and incorrectly tagged words are 28. From Anaemia 735 words are correctly tagged and 58 words are incorrectly tagged. From Migraine 130 correctly tagged words and 30 incorrectly tagged words. And from Keloids 810 words correctly tagged and 110 words incorrectly tagged. Hence total tagged words are 2322 out of which 2096 are correctly tagged and 226 are incorrectly tagged. The accuracy of POS tagging is revealed in the table 3.

Table 3: Accuracy of POS Tagging

| *Diseases(from corpus)* | *Accuracy (%) of Correctly tagged words* |
|---|---|
| Rheumatism | 93.76 % |
| Anaemia | 92.68 % |
| Migraine | 81.25 % |
| Keloids | 88.04 % |
| Average accuracy | 88.93 % |

From table 3. Accuracy of correctly tagged words from Rheumatism is 93.76%, Anaemia is 92.68%, Migraine is 81.25%, and Keloids is 88.93%.Total accuracy is 88.93% was achieved by the proposed tagger. Whereas dTagger had accuracy of 87%. So, proposed clinical tagger had much better accuracy than dTagger.

## 4. Conclusions

In clinical domain this is the foremost time that homoeopathy sentences were tagged. The proposed Part of Speech tagger of homoeopathy was developed manually. The resulting accuracy was computed to 88.93%. We use untagged Homoeopathic corpus of 20085 words, corpus is categories into different diseases. We computed correctly and incorrectly tagged words 2096 and 226 respectively. For tagging we had assembled four diseases (Rheumatism, Anaemia, Migraine, and Keloids). Sentences of each disease were autonomously tagged with accuracy 93.76%, 92.68%, 81.25%, and 88.04%, respectively, and the average percentage is computed to 88.93%. To acquire higher accuracy, hefty data is required. In addition to that, data should be taken from special homoeopathy books, patient's medical report and symptoms of different diseases. We plan to broaden the homoeopathy corpus up to 170,000 words.

**Dr. S.K. Dwivedi** has obtained his Ph.D. Degree from Banasthali Vidyapeeth in the year 2006. He has completed his Ph.D. in the area of Web Mining. His research interest are Web content Mining, Semantic Web, Search Engine performance evaluation etc. He has published many of the valuable research papers in various national and international Journals. He is presently working as a Associate Professor of Computer Science dept, of BBAU, Lucknow, India.

**Pramod P. Sukhadeve** has obtained his MSc. Degree in the year 2006 from Nagpur University. His research interest are Natural Language Processing, Machine Translation System and in Homoeopathy. He has published some of the research papers in refereed Journals and international conferences. Recently he is doing full time research from BBA University (A CCENTRAL UNIVERSITY) Lucknow,